# Fitness-based Adaptive Control of Parameters in Genetic Programming

Adaptive Value Setting of Mutation Rate and Flood Mechanisms

Michal Gregor
Department of Control and Information Systems
Faculty of Electrical Engineering, University of Žilina
Žilina, Slovak Republic
e-mail: o.m.gregor@gmail.com

Juraj Spalek
Department of Control and Information Systems
Faculty of Electrical Engineering, University of Žilina
Žilina, Slovak Republic
e-mail: juraj.spalek@fel.uniza.sk

*Abstract*—This paper concerns applications of genetic algorithms and genetic programming to tasks for which it is difficult to find a representation that does not map to a highly complex and discontinuous fitness landscape. In such cases the standard algorithm is prone to getting trapped in local extremes. The paper proposes several adaptive mechanisms that are useful in preventing the search from getting trapped.

*Keywords-genetic algorithms; genetic programming; adaptive mechanism; adaptive parameter setting*

## I. INTRODUCTION

Genetic algorithms and genetic programming represent a well known optimization method. Among their strengths is the flexibility of representation, which allows for their application to a wide variety of tasks. Even though the representation of the solution is not forced to take a single prescribed form, it is still required that it follow certain guidelines so that the algorithm is able to search the resulting fitness landscape with reasonable efficiency.

In certain tasks, such as evolving algorithms with memory, it may be very difficult to find a representation that does not map to a highly complex and discontinuous fitness landscape. This makes the process of search prone to getting trapped in local extremes for such tasks.

This paper presents several adaptive mechanisms that aim to improve properties of the standard genetic programming with respect to this problem. These are all based on the observation that it is often possible to help the algorithm escape from local maxima by introducing new genetic material into the process. There are multiple ways to achieve this, some of which will be presented in this paper.

Following sections present respectively – the model problem to which the adaptive mechanisms were applied; an overview of existing approaches to parameter control; and description of the proposed adaptive mechanisms as well as the results.

## II. THE MODEL PROBLEM

Let us first briefly describe the model problem on which the results achieved by various versions of the proposed adaptive mechanisms are to be presented – *the artificial ant problem*. Application of genetic programming to this task has been described in detail by John Koza [1].

The artificial ant problem is essentially a trail-following task. The actor – an artificial ant is to navigate in an environment, following an irregular path consisting of pieces of food which it is supposed to collect. The ant has severely limited sensing capabilities – it only sees a single tile that is right in front of it.

The solution is represented by a simple syntactic tree. Automatically defined functions, recursion and other advanced concepts are not utilized.

In contrast to Koza's original application, where the syntactic tree represents a controller (i.e. the program runs, orders the agent to perform a certain action, waits for its completion and then continues to run the same way until the end of the program is reached at which point the whole syntactic tree is re-executed), the mode of execution has been modified in our implementation – the whole syntactic tree is now run in every step and it is used to determine which action the agent should take. This means that some explicit memory model, such as indexed memory, is now required to make account of previous inputs and actions.

This approach has several advantages, but it also makes the solution considerably more difficult to evolve. Some of the underlying issues have been discussed by Astro Teller in [2]. We shall, at this point, confine ourselves to concluding that standard genetic programming has shown itself to be unable to solve the model problem in the prescribed mode (that is, none of the solutions has achieved the maximum fitness score). The search easily becomes trapped in local extremes of the fitness landscape.

The parameters of the algorithm are as follows: the maximum of 150 generations; 500 individuals; the maximum depth of the tree set to 10; the Santa Fe trail (Fig. 1).

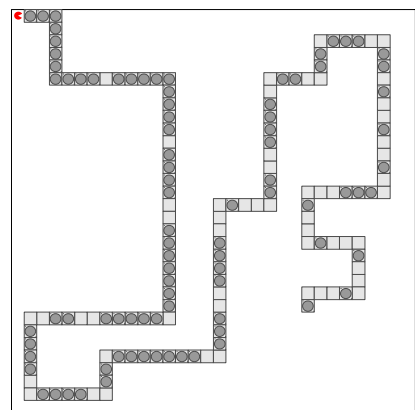

Figure 1. The Santa Fe trail

## III. Existing Approaches to Parameter Control

As mentioned, in some applications based on the theory of genetic algorithms and genetic programming, the fitness landscape can be so complex that additional techniques may be required in order to find the global optimum.

Among the approaches that aim to enhance various properties of the standard algorithm, such as convergence speed and resilience to getting trapped in local extremes, are different parameter setting schemes. There are many different approaches most of which would fall into one of the following categories [3], [4]:

- static parameter control,
- dynamic parameter control,
- adaptive parameter control,
- self-adaptive parameter control.

### A. Static Parameter Control

The common feature of static parameter control approaches is that the setting they provide remains constant for the entire duration of the evolutionary process. There are many works analyzing the problem of finding optimum settings for parameters like mutation rate and crossover rate. Some of these are listed in [3], e.g. the work of Mühlenbein, which proposes the following formula for the mutation rate:

$$p_m = 1/L, \quad (1)$$

where $L$ is the length of the bit string by which the individual is represented.

### B. Dynamic Parameter Control

As stated in [4], dynamic parameter approaches typically prescribe a deterministically decreasing schedule over a number of generations. The following formula for mutation rate derived by Fogarty is also provided:

$$p_m(t) = \frac{1}{240} + \frac{0.11375}{2^t}, \quad (2)$$

where $t$ is the generation counter.

Articles [3], [4] also both refer to a more general expression derived by Hesser and Männer:

$$p_m(t) = \sqrt{\frac{\alpha}{\beta}} \times \frac{\exp\left(\frac{-\lambda t}{2}\right)}{\lambda \sqrt{L}}, \quad (3)$$

where $\alpha$, $\beta$, $\gamma$ are constants, $\lambda$ is the population size and $t$ is the generation counter. $L$ is the length of the bit string.

### C. Adaptive Parameter Control

Adaptive parameter control techniques monitor the search process itself and provide feedback. Some examples can be found in [5]. The authors propose the following formulas for crossover and mutation probability respectively:

$$p_c = \begin{cases} k_1 \dfrac{f_{max} - f'}{f_{max} - \bar{f}} & f' > \bar{f} \\ k_3 & f' \leq \bar{f} \end{cases}, \quad (4)$$

$$p_m = \begin{cases} k_2 \dfrac{f_{max} - f}{f_{max} - \bar{f}} & f > \bar{f} \\ k_4 & f \leq \bar{f} \end{cases}, \quad (5)$$

where $f$ is the fitness value of the individual to be mutated, $f'$ is the larger of the fitness values of the individuals to be crossed and $k_3$ and $k_4$ are constants. It is required that $k_1$ and $k_2$ be less than 1.0 in order to constrain $p_c$ and $p_m$ to the range of $\langle 0,1 \rangle$. The $p_c = k_3 \quad f' \leq \bar{f}$ and $p_m = k_4 \quad f \leq \bar{f}$ expressions are to prevent crossover and mutation probabilities from exceeding 1.0 for suboptimal solutions.

The authors call this approach AGA (Adaptive Genetic Algorithm). Paper [5] also observes that $p_c$ and $p_m$ are zero for the solution with maximum fitness and that $p_c = k_1$ for $f' = \bar{f}$, while $p_m = k_2$ for $f = \bar{f}$. For further details and for information concerning setting the values of the constants refer to [5].

### D. Self-adaptive Parameter Control

When using the self-adaptive parameter control approach, parameters such as mutation rate and crossover probability of each individual are part of its genome and are evolved with it. As stated in [4], the idea behind this is that a good parameter value will provide an evolutionary advantage to the individual. For further reference see [3] or [4].

## IV. Adaptive Value-switching of Mutation Rate

Most of the existing parameter setting mechanisms, as presented in the previous section, either focus on setting GA-specific parameters such as length of the bit string (e.g. rule (1)), or are not adaptive (e.g. (1), (2) and (3)). The AGA mechanism described in [5] behaves adaptively, but its purpose is to speed up convergence, which (as shown later) makes the problem worse as AGA does not discern between local and global optima and thus effectively increases the probability of getting trapped.

Furthermore, it is obvious that equations (4) and (5) assign the best individual zero crossover and mutation probabilities, while assigning high probabilities to less fit individuals. The reasoning behind this is that the less fit individuals can safely be disrupted by high mutation rates and recombined by crossover, thus employing the solutions with subaverage fitness to search the space of solutions [5], while highly fit individuals are preserved.

However, such approach has a very obvious downside which does not seem to be addressed – the highly fit individuals obviously contain the most excellent genetic material available and by disallowing mutation and crossover for these individuals the genetic code they carry effectively becomes isolated and is not used to generate new solutions.

## A. Description of the AVSMR Mechanism

The idea that the most fit solutions should survive crossover and mutation unmodified is obviously valid, yet that feature can be enforced by using elitism (i.e. the best individual is copied to the next generation unmodified). Keeping that in mind we propose a new adaptation scheme called AVSMR: *Adaptive Value-switching of Mutation Rate*. The main idea is that the mutation rate should be increased to a high value when the search has become trapped in an extreme so as to provide the search process with new genetic material some of which may previously have been unavailable. To determine whether the search has become trapped the adaptive mechanism observes the change of average fitness in time.

To describe the solution in more detail – the algorithm works with 2 values of mutation probability: the normal value and the high value. The algorithm switches from the normal value to the high value once the trigger criterion activates.

The trigger criterion itself is based on a measure that we will herein term a *delta sum*:

$$\Delta S_i = \alpha \cdot \Delta S_{i-1} + \frac{\bar{f}_i - \bar{f}_{i-1}}{\bar{f}_i}, \qquad (6)$$

where $\Delta S_i$ is the delta sum in generation $i$; $\bar{f}_i$ is the average fitness in generation $i$ and $\alpha$ is the feedback coefficient (the experiments have been carried out for $\alpha = 0.4$).

If the delta sum is lower than a preset value for a predefined number of generations, that is to say the increase of average fitness in the last few generations is low, indicating that the search has become trapped – the mutation rate is set to its high value so as to provide the search with new genetic material. As mentioned before, when used in conjunction with elitism it is guaranteed that the best solution is not destroyed by the high mutation probability.

The mutation rate is reset back to its normal value when at least one of the following conditions is true:
- the average fitness increases enough to produce a sufficiently large delta sum;
- the maximum fitness increases;
- mutation has been set to its high value for at least $n$ generations.

The *n*-generation limit is to ensure that the activation does not go on indefinitely (with the high mutation probability it is not very likely that the average fitness will increase sufficiently to satisfy the first condition and maximum fitness may not increase as well).

It has been observed that average fitness typically decreases when the criterion activates because the search process is to a large extent disrupted by the high mutation rate. However after the *n*-generation limit forces the mutation rate back to its normal value, average fitness tends to increase rapidly, thus usually moving away from the local extreme.

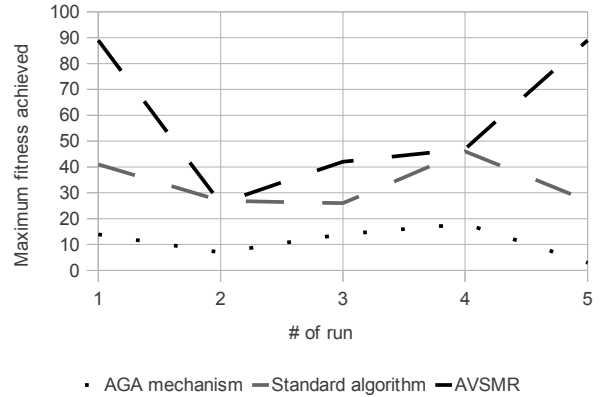

Figure 2.  Comparison of AGA and AVSMR

## B. Experimental Results

Several experiments have been carried out. Fig. 2 shows performance of the search algorithm with the AGA adaptive mechanism proposed in [5] (with constants set according to recommendations). It also shows performance of the search algorithm without any adaptive mechanism and with the AVSMR mechanism proposed in this paper. The maximum fitness value achieved is shown for each of the 5 runs that are displayed.

As shown, search achieves suboptimal results when running with no adaptive mechanism. This can be ascribed to its inability to escape from local extremes. With no adaptive mechanism the search has not found the global optimum (fitness = 89) in any of the 5 runs.

As mentioned, the AGA mechanism causes further deterioration and its results are thus even worse than those produced by the standard algorithm.

The AVSMR mechanism proposed in this work improves the process of search – in 2 of the runs the global optimum is found, yet in certain cases not even the high mutation rate is guaranteed to help the search to escape from local maxima (runs 2, 3, 4).

## V. THE SIMPLE FLOOD MECHANISM

The AVSMR mechanism described in previous section is helpful in controlling the search process by helping it to escape from local extremes, yet it is not completely reliable and not always effective. To address these issues, we have developed another adaptive scheme supposed to provide new genetic material to an even larger extent.

## A. Simple Flood Mechanism

The principle is very straight-forward – once a trapping is detected – a relatively small part of the population is selected: these individuals survive. The rest of the population is destroyed and replaced by newly generated individuals. This method is superior to AVSMR in that a large part of the population is guaranteed to be replaced and the newly generated individuals can be (and need not be) generated in the same way that the initial population was.

The trigger criterion has also been modified. The first requirement is that the criterion only activates for a single generation at a time as it would probably be useless and possibly even counterproductive to activate the flood mechanism for several successive generations.

The new trigger criterion is still based on the average fitness $\bar{f}_i$ (where $i$ is the number of generation). The criterion stores average fitness $\bar{f}_i$ for $N$ generations (that is, $N-1$ previous generations and the current one; $N=7$ generations was used in the experiments). The mechanism cannot activate before $\bar{f}_i$ for at least $N$ generations has been collected. Once that is true, the mechanism activates if the following holds:

$$\sum_{i=j}^{j-(N-2)} \bar{f}_i - \bar{f}_{i-1} < \Theta, \qquad (7)$$

where $j$ is the number of current generation and $\Theta$ is an activation threshold. It is also possible to interpret the threshold as a relative parameter in which case we can rewrite the equation as follows:

$$\frac{\sum_{i=j}^{j-(N-2)} \bar{f}_i - \bar{f}_{i-1}}{\bar{f}_i} < \Theta. \qquad (8)$$

All experiments were carried out using (8).

It is also important to note that once the mechanism activates, the array storing the previous values of average fitness is cleared so it is guaranteed that the mechanism does not activate for next $N$ generations.

Although the approach seems straight-forward and similar in concept to AVSMR, experimental results point out an important issue. As obvious from Fig. 3, the results achieved by the Simple Flood Mechanism are significantly worse than those produced by AVSMR – they are in fact worse than those produced by the standard algorithm.

The reason behind this is very simple – although we do introduce new genetic material into the process, the newly generated individuals will generally have very low fitness (usually 0, 3, or 4 at most). Therefore if we apply fitness-proportionate selection to these in the next generation, almost every newly generated individual will be discarded. The survivors on the other hand will now dominate the population. This is especially true later in the evolutionary process when fitness score of the best individual will tend to be vastly greater than that of any randomly generated individual. At this point the next generation will be formed almost exclusively by the best individual, which will nearly in every case aggravate the problem of getting trapped in local extremes instead of solving it.

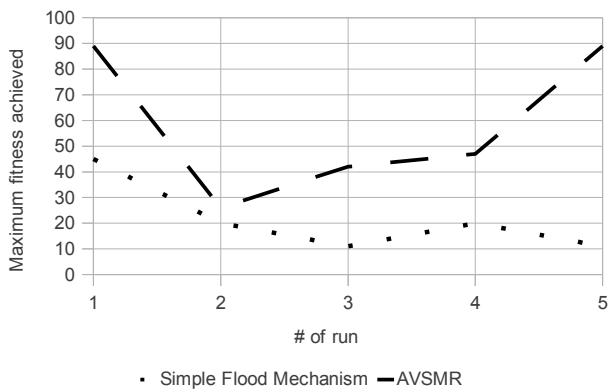

Figure 3. Comparison of the AVSMR and the Simple Flood Mechanism

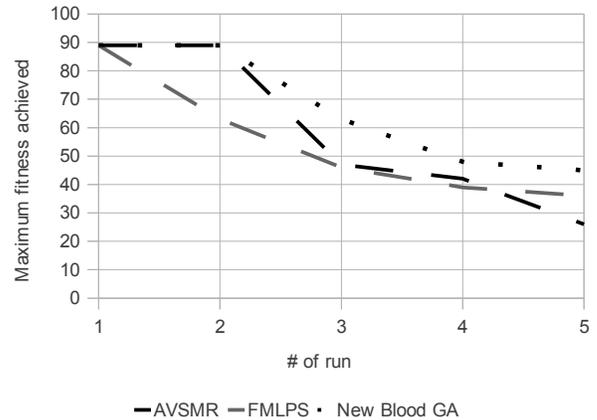

Figure 4. AVSMR, FMLPS and the New Blood Mechanism

### B. Flood Mechanism with Low-pressure Scaling and the New-Blood Mechanism

There are several ways to alleviate the problem that the Simple Flood Mechanism faces. The objective is – in any case – to create such scheme in which the newly generated individuals mate with the survivors so as to make use of their potentially useful code.

This paper proposes two different ways to achieve this:
- apply a fitness scaling function with low selection pressure to the GA for several generations following the flood – this mechanism will be referred to as *Flood Mechanism with Low-pressure Scaling (FMLPS)*;
- once the mechanism activates, create only such mating pairs in which at least one individual is newly generated – this mechanism will be referred to as the *New Blood Mechanism*.

The experimental results are shown in Fig. 4. FMLPS uses power scaling of 0.3 as the low-pressure scaling. To make the comparison easier, the values are now ordered by fitness rather than by the number of run. This shows that AVSMR is still superior to FMLPS (although FMLPS is – in contrast to the Simple Flood Mechanism presented earlier – significantly better than standard GP). The New Blood GA on the other hand is definitely superior to AVSMR – although it still gets trapped in local extremes, the maximum fitness values achieved are generally greater than those achieved by the AVSMR.

### VI. SUGGESTIONS FOR FURTHER WORK

It has been shown that the adaptive mechanisms described in this work are able to effect considerable improvements. They are able to prevent the search algorithm from getting trapped in local maxima to a certain extent. Further experiments should now be carried out to test usefulness of these approaches in a wider range of applications for which genetic algorithms and genetic programming are prone to getting trapped in local extremes.

It has also become apparent that neither with the high mutation rates, nor with various versions of the flood mechanism it is always possible to guarantee success. Value-switching, or even piecewise continuous relationships for other parameters could perhaps help to alleviate the problem to a further extent. The trigger criteria would also require some additional work in order to become more stable and robust.

### VII. CONCLUSION

It has been shown that genetic algorithms and genetic programming may be applied to such tasks in which it is difficult to find a representation that does not map into a highly complex fitness landscape. Evolving algorithms with explicit memory concepts, such as indexed memory, may be considered an example of such task as shown by Teller in [2] and demonstrated on the model problem in this paper.

Tasks with highly complex and discontinuous fitness landscapes are most difficult to solve using standard genetic algorithms and genetic programming as both of these methods are likely to get trapped in local extremes when applied to such problems.

The paper presents several adaptive mechanisms – the AVSMR mechanism, the FMLPS mechanism and the New Blood GA. These mechanism are shown to be able to alleviate the problem to a certain extent, thus achieving significantly better results that the standard algorithm.

Although the results compare favorably to those of the standard algorithm, even the proposed mechanisms cannot always guarantee that the search process will indeed escape from every local extreme that it encounters.

Related techniques such as adaptive value-switching or piecewise continuous relationships for other parameters of the search algorithm might provide further improvements. The influence that some of the flood mechanism related parameters (such as the number of survivors, or the selection pressure applied by the low-pressure scaling) have on the process of search may also provide area for further research.